\documentclass[runningheads]{llncs}
\let\llncssubparagraph\subparagraph
\let\subparagraph\paragraph
\usepackage[compact]{titlesec}

\usepackage{graphicx}
\usepackage{amsmath,amssymb} 
\usepackage{titlesec}

\let\subparagraph\llncssubparagraph
\usepackage{enumitem}

\usepackage{booktabs} 

\usepackage{amsmath,bm}
\usepackage{amsfonts}
\usepackage{amssymb}
\usepackage{multirow}
\usepackage{flushend}
\usepackage[boxruled, linesnumbered]{algorithm2e}
\usepackage{bbold}
\usepackage[font=small,belowskip=-7pt,aboveskip=0pt]{caption}

\titlespacing\section{0pt}{8pt plus 2pt minus 1pt}{6pt plus 2pt minus 1pt}
\titlespacing\subsection{0pt}{8pt plus 2pt minus 1pt}{4pt plus 2pt minus 1pt}

\begin{document}

\title{Enhancing Perceptual Attributes \\ with Bayesian Style Generation}
\titlerunning{Bayesian Style Generation} 


\author{Aliaksandr Siarohin\inst{1}\orcidID{0000-0001-9252-1775} \and
Gloria Zen\inst{1}\orcidID{0000-0001-6271-5915} \and
Nicu Sebe\inst{1}\orcidID{0000-0002-6597-7248} \and
Elisa Ricci\inst{1,2}\orcidID{0000-0002-0228-1147}}

%

\authorrunning{A. Siarohin et al.} 


 \institute{DISI, University of Trento, Italy 
 \email{\{aliaksandr.siarohin,gloria.zen,e.ricci,niculae.sebe\}@unitn.it}\\
 \and
 Fondazione Bruno Kessler (FBK), Trento, Italy
 }

\maketitle

\begin{abstract}
Deep learning has brought an unprecedented progress in computer vision 
and significant advances have been made in predicting subjective properties inherent to visual data (e.g., memorability, aesthetic quality, evoked emotions, etc.). 
Recently, some research works have even proposed deep learning approaches to modify images such as to appropriately alter these properties. 
Following this research line, this paper introduces a novel deep learning framework for synthesizing images in order to enhance a predefined perceptual attribute. Our approach takes as input a natural image and exploits recent models for deep style transfer and generative adversarial networks to change its style in order to modify a specific high-level attribute. 
Differently from previous works focusing on enhancing a specific property of a visual content, we propose a general framework and demonstrate its effectiveness in two use cases, \textit{i.e.} increasing image memorability and generating scary pictures. 
We evaluate the proposed approach on publicly available benchmarks, demonstrating its advantages over state of the art methods.

\end{abstract}

\section{Introduction}
\label{sec:intro}
The recent advances in predicting and understanding subjective properties of visual data (\textit{e.g.} beauty, memorability, interestingness, etc.) enabled by deep learning models \cite{lu2015deep,mai2016composition,gygli2013interestingness,khosla2015understanding,alameda2017viraliency,porzi2015predicting} have motivated researchers in computer vision to take a step forward and investigate automatic techniques to manipulate images in order to modify these properties. For instance, recent works have proposed methods to edit images in order to increase their memorability \cite{siarohin2017make}, to improve their aesthetic quality \cite{wang2017deep} or to evoke specific emotional reactions into users \cite{peng2015mixed}.
Recently, deep style transfer methods  \cite{huang2017arbitrary,gatys2016image,li2017universal,ulyanov2016texture} which allow the users to modify pictures by blending them with style images have gained popularity. These methods have significantly widened the 
set of editing 
operations available in traditional image enhancement tools, fostering the diffusion of novel 
software for turning user pictures into artworks.
While earlier methods for neural style transfer \cite{gatys2016image,ulyanov2016texture} considered a fixed set of styles and relied on slow optimization processes, more recent approaches \cite{huang2017arbitrary,li2017universal} 
are highly flexible, enable the generation of arbitrary styles and have close to realtime performance.

\begin{figure}[t]
\centering
\includegraphics[width=\linewidth]{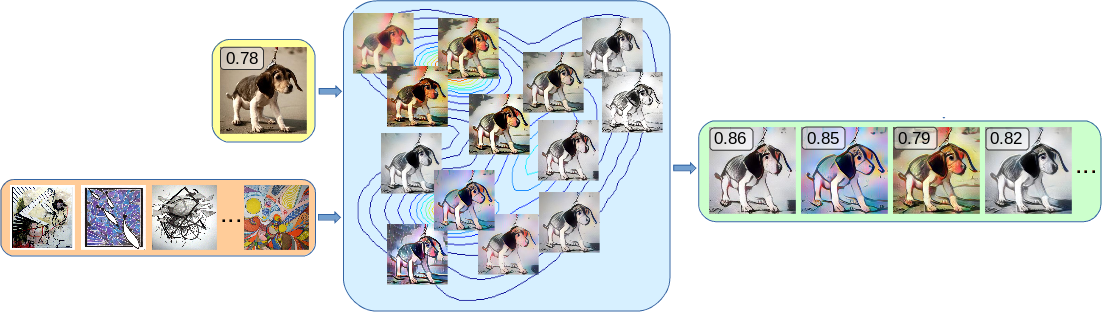}
\caption{Idea behind our approach. 
Given a generic input image (yellow box) our framework provides as output a set of stylized images (green box) obtained by applying the styles which maximally enhance a given perceptual attribute. The attribute value (shown on top left corners of input and output images) is automatically assessed by a deep network.
The style selection process is achieved by modeling the style space (light blue box) as a probability distribution automatically learned from a given training set of style images (orange box) using a generative adversarial network.
}
\label{fig:teaser}
\end{figure}

Motivated by these recent advances, in this paper we propose a novel approach
for generating stylized images in order to enhance a given perceptual attribute. Similarly to previous deep style transfer methods \cite{huang2017arbitrary,li2017universal}, the stylized images are obtained by training a feed-forward neural network which receives as input the original images and the style pictures. Opposite to previous works,
the style choice is not made by a user but it is automatic and is driven by a specific criterion, \textit{i.e.} increasing the value of the given perceptual attribute. 
At the core of our style selection process there is a novel probabilistic framework which exploits recent Generative Adversarial Networks (GANs) to learn a probability distribution modeling the style space and Markov Chain Monte Carlo (MCMC) methods to sample from the learned distribution and compute the best styles. We named the proposed approach BAE (Bayesian Attribute Enhancement). 
While our framework is generic and can be used for different types of perceptual attributes, in this work we focus on two applications, \textit{i.e.} increasing memorability, defined as the probability of an image to be remembered \cite{khosla2015understanding} and generating scary pictures.
We quantitatively and qualitatively evaluate the proposed approach on publicly available datasets, demonstrating superior performance over state of the art methods. 
Figure~\ref{fig:teaser} illustrates the intuition behind our method. 

\textbf{Contributions.} 
To summarize, the contribution of this work is threefold. (i)
    We propose a novel framework to automatically modify an input image in order to alter its inherent perceptual attributes. To preserve the semantic content of the original image, our approach relies on a neural style transfer method. In this way, the problem of perceptual attribute enhancement naturally translates to that of retrieving the best styles to apply to the given image. Opposite to previous works which focus on modifying a specific subjective property \cite{siarohin2017make,wang2017deep,peng2015mixed}, our method can be applied to any arbitrary attribute. While we tested it on two scenarios, we expect the method to be useful in other applications, \textit{e.g.} for enhancing the aesthetic quality of images or for increasing their virality score.
(ii) By exploiting state of the art deep style transfer techniques \cite{huang2017arbitrary} within a novel probabilistic framework for modelling the style space, our approach does not simply select the best styles from a small predefined set but also allows to generate arbitrary new styles. Thus, a higher increase of the attribute score can be obtained with respect to previous approaches \cite{siarohin2017make}. 
(iii) Our framework is highly flexible and allows not only to automatically select the best styles 
    but also the degree of stylization. 
Furthermore, by resorting on MCMC sampling methods, it can be used to compute multiple styles. In this way we keep the users in the loop, suggesting the best styles for attribute increase but still allowing the users to choose among multiple stylized images 
according to their personal preferences. 






\section{Related Works}
Our work lies at the intersection between two main research lines. The first line focuses on the problem of understanding and predicting subjective properties from visual data, 
the second one includes works proposing novel deep models for automatic image editing. 

\textbf{Predicting perceptual attributes from visual data.}
In the last decade several works in computer vision and multimedia have addressed the problem of modelling and predicting perceptual attributes from images and videos. These studies have focused on the automatic assessment of aesthetic value \cite{lu2015deep,mai2016composition},
interestingness \cite{gygli2013interestingness},
memorability \cite{khosla2015understanding},
virality \cite{alameda2017viraliency},
symmetry \cite{Funk_2017_ICCV}, etc. In some cases, typically where a large amount of training data is available, automatic systems can even reach human-level performances. 
For instance, Khosla \textit{et al.} \cite{khosla2015understanding} showed that a deep learning model trained on LaMem, the largest memorability dataset so far, can predict image memorability with an accuracy close to that of human annotators.
Similarly, recent methods for computing automatically the aesthetic value of images are quite precise, achieving an accuracy superior to 75\% on the AVA dataset \cite{lu2015deep,mai2016composition,peng2016toward}. 
In this work we focus not only on predicting subjective attributes but we also address the more challenging task of image enhancement.

\textbf{Deep Models for Automatic Image Manipulation.}
Deep learning models and, in particular, neural style transfer methods \cite{huang2017arbitrary,gatys2016image,li2017universal,ulyanov2016texture} and deep generative networks \cite{goodfellow2014generative,isola2017image} have enabled significant advances for automatic image editing and generation. In the wake of these progresses, recent works have taken a step beyond perceptual attributes prediction and have proposed methods to manipulate images in order to modify these intrinsic attributes \cite{wang2017deep,siarohin2017make,tsai2017deep}.  
For instance, Wang \textit{et al.} \cite{wang2017deep} addressed the task of increasing the aesthetic value of an image by finding the best crop. 
Tsai \textit{et al.} \cite{tsai2017deep} proposed a deep model for image harmonization which adjusts the appearance of the image foreground 
in order to better adapt it to the background.
Liao \textit{et al.} \cite{liao2017visual} introduced a method to alter intrinsic image properties like color, texture or style based on deep analogy and visual property transfer.
However, these works simply propose strategies to modify a specific property of images but 
do not provide a general framework to systematically enhance an arbitrary perceptual attribute and quantitatively assess its value increase. Recently, Siarohin \textit{et al.} \cite{siarohin2017make} moved a step forward in this direction, by proposing an approach which selects the best styles for a given image in order to increase its memorability. Still, their method 
relies on a pre-defined set of styles and the degree of stylization is also fixed \textit{a priori}. In this work, we overcome these limitations by introducing a more general and flexible approach which operates on a large set of styles and where the trade-off between style and content is regulated by a user-defined hyper-parameter $\alpha$. 

\section{Enhancing Perceptual Attributes with BAE}
\label{sec:method}

As stated in Section \ref{sec:intro}, the proposed approach deals with the problem of automatically modifying an arbitrary input image in order to enhance a specific perceptual attribute, \textit{e.g.} its memorability, the likelihood to evoke specific emotional reactions from users, etc. This task is addressed within a novel Bayesian framework and by resorting on a state of the art neural style transfer method \cite{huang2017arbitrary}. In fact, our approach aims to modify the given image increasing its perceptual attribute score by changing its style while retaining the semantic content. 
In the following we briefly describe the neural style transfer method in \cite{huang2017arbitrary} and then introduce the proposed approach providing some details on our implementation.

\subsection{Arbitrary Style Transfer}
\label{sec:arbitrary}
Given an input image $I$ and a style image $S$, let us denote with $\hat{I} = {T(I, S)}$ the modified image obtained by applying the style transfer model $T$. In this work we consider the style transfer approach in \cite{huang2017arbitrary} as, oppositely to earlier methods \cite{gatys2016image,ulyanov2016texture}, i) it is not tied to a fixed set of styles, allowing to generate arbitrary new styles, ii) it performs style transfer in realtime, and iii) it is very flexible, enabling to control the degree of stylization also at test time.

The deep architecture $T$ proposed in \cite{huang2017arbitrary} has a simple encoder-decoder structure. The encoder $f_E$ is used to compute the feature maps $f_E(I)$ and $f_E(S)$ associated respectively to the input and to the style images. The computed feature maps are then fed to a specific feature alignment layer, the Adaptive Instance Normalization (AdaIN) layer. This layer aligns the mean $\mu$ and variance $\sigma$ of the image features to those of the style features, producing the target feature maps:
\begin{equation}
    t = \mbox{AdaIN}(f_E(I),f_E(S))=\sigma (f_E(S)) \nu(f_E(I))+\mu(f_E(S)) \label{eq:adain}
\end{equation}
where $\nu (x)= \frac{x-\mu(x)}{\sigma(x)} \label{eq:norm}$.
The decoder $f_D(t)$ is trained to map the target feature maps $t$ 
back to the image space, generating the stylized image $\hat{I}=T(I,S)=f_D(t)$.
As typically done in neural style transfer methods, the network $T$ is trained by optimizing a loss function which is the weighted sum of two terms, \textit{i.e.} $\mathcal{L}=\mathcal{L}_c+\gamma \mathcal{L}_S$, where $\mathcal{L}_c$ and $\mathcal{L}_S$ are the content and the style loss respectively and $\gamma$ is a user defined parameter regulating at training time the trade-off between semantic content and stylization. We refer the reader to the original paper \cite{huang2017arbitrary} for details on the definition of the loss functions.

A prominent feature of the neural style transfer method in \cite{huang2017arbitrary} is the possibility to control the degree of stylization not only at training time by changing $\gamma$ but also at test time. In particular, a parameter $\alpha$ is introduced and the stylized image is computed as:
\begin{equation}
    \hat{I}=T(I, S, \alpha)=f_D(\alpha f_E(I) + (1-\alpha) t) \label{eq:alpha}
\end{equation}
Here, $\alpha=1$ corresponds to the case where no style transfer is performed, while $\alpha=0$ corresponds to full stylization.

\subsection{Bayesian Attribute Enhancement}
\label{sec:BAE}
The main idea behind the proposed approach is to construct a model which, given an arbitrary image $I$ and a set of style images $\Omega_S=\{S_i\}_{i=1}^K$, is able to automatically compute a novel set of $M$ styles that, applied to the input image, fully enhance a specific perceptual attribute, \textit{e.g.} increase its memorability. 
To build this model, inspired by \cite{huang2017arbitrary}, we first introduce a compact representation for styles in terms of mean and standard deviation of activations. Formally, given a style image $S$ and a pre-trained encoder network $f_E$, we define a style $s = \{\mu_s, \sigma_s\}$ where $\mu_s=\mu(f_E(S))$ and $\sigma_s=\sigma(f_E(S))$. 

\begin{figure}[t]
\centering
\includegraphics[width=0.7\linewidth]{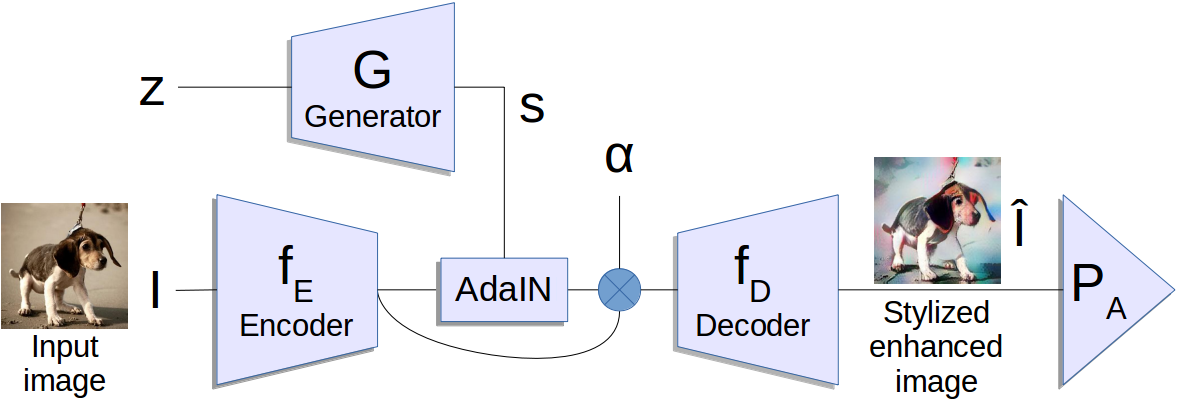}
\caption{Overview of our approach: given an input image $I$ our method generates through $G$ the style $s$ which maximizes the perceptual attribute score, computed by $P_A$, of the stylized image $\hat{I}$. }
\label{fig:flow}
\end{figure}

Given the set of style images $\Omega_S=\{S_i\}_{i=1}^K$ and the associated representations $\mathcal{S}=\{s_i\}_{i=1}^K$, we propose to learn a probability density function $P_{\mathcal{S}}(s)$ modelling the style space. 
While different methods can be used for this purpose, motivated by the recent successes of deep generative models \cite{goodfellow2014generative}, in this paper we consider a Generative Adversarial Network (GAN).
A GAN consists of two networks, a generator $G$ and a discriminator $D$. These two networks play a minimax game in which the task of $D$ is to distinguish the samples generated by $G$ from the real samples and the task of $G$ is to increase the chances of $D$ producing a high probability for a synthetic example. In \cite{goodfellow2014generative} it is shown that the equilibrium in this game is achieved when the probability density of the generated samples is equal to the probability density of the real ones. For our application we use $\mathcal{S}$ as real samples, and learn the generator $G$ in order to produce styles ${s = G(z)}$, where $z \sim P(z)$. The input $z$ to the generator is sampled from some simple noise distribution $P(z)$ such as a Gaussian distribution $N(0, \mathbb{1})$. 
For training the GAN model we use an efficient version of Wasserstein GANs \cite{arjovsky2017wasserstein}, and specifically WGAN-GP, recently proposed in \cite{gulrajani2017improved}.

In addition to the GAN model, we propose to learn two additional deep networks. The first network implements the neural style transfer approach described in Section~\ref{sec:arbitrary}. In the following, given a style $s$ we denote as $\hat{I}=Z(I,s)=f_D\left(\sigma_s \nu(f_E(I))+\mu_s\right)$ the stylized image.
The second network is used to learn a function $P_A(I)$ which, given an input image $I$, outputs a probability score reflecting the strength of a given perceptual attribute. The design of this network and its training strategy, described in Section~\ref{sec:implem}, is at the core of our method and depends on the chosen subjective attribute. In particular in this paper we consider two attributes, memorability and scariness, \textit{i.e.} we propose two different criteria for modifying pictures: increasing their memorability and maximizing their likelihood to evoke scary reactions into users. 
Given the above definitions, we propose to build a probability density for the joint model: 
\begin{eqnarray}
 &  P(\hat{I}, z) & = P(\hat{I}|z)P(z)  = P_A(Z(I, G(z))) N(z; 0, \mathbb{1})  \label{eq:probz} 
\end{eqnarray}
where the last term is derived 
considering the learned models $G$, $Z$ and $P_A$. 
In this way, we obtain a probability over $z$ which, in our work, can be seen as a latent representation of a style $s$. 
We propose to exploit $P(\hat{I}, z)$ in order to find the styles which better enhance a given perceptual attribute. 
Specifically, to obtain a diverse set of styles corresponding to high values of the target attribute, we propose to sample from $P(\hat{I}, s, z)$ using MCMC methods. The best styles, in fact,
correspond to the modes of the distribution.

We also extend the proposed Bayesian framework in order to compute automatically not only the best styles but also the degree of stylization. To this aim, we consider Eqn.~\ref{eq:alpha} and define $\hat{Z}(I, S, \alpha)=f_D(\alpha f_E(I) + (1-\alpha) t)$. However, instead of setting $\alpha$ as a constant, we assume that $\alpha$ is a random variable. In this case, similarly to Eqn.~\ref{eq:probz}, we define the joint probability: 
\begin{equation}
   {P(\hat{I}, z, \alpha)} = P(\hat{I}|z,\alpha)P(z)P(\alpha)
\end{equation}
where $P(\hat{I}|z,\alpha)=P_A(\hat{Z}(I, G(z),\alpha))$ and $P(\alpha)$ is a prior probability. 
In this case with MCMC sampling we obtain a set of latent style representations  $\mathcal{S}'=\{z_1, z_2, ... z_M\}$, 
as well as a set of stylization coefficients $\mathcal{A}'=\{\alpha_1, \alpha_2, ... \alpha_M\}$.
In the following, we refer to our method as Bayesian Attribute Enhancer (BAE), while its adaptive version where we also automatically compute $\alpha$ value is called ABAE. An overview of the proposed framework is illustrated in Fig.\ref{fig:flow}.


\subsection{Implementation}
\label{sec:implem}
In this Section we report additional details on the implementation of the proposed method. In particular, we describe the adopted deep network architectures and provide further details on the considered MCMC sampling strategies.

\textbf{Network Architectures.} 
The neural style transfer network is implemented following the original paper \cite{huang2017arbitrary}. The encoder $f_E$ is built from the first four convolutional layers of a pre-trained VGG-19 \cite{simonyan2014very}. The decoder $f_D$ is implemented with a structure mirroring the encoder, with all pooling
layers replaced with up-sampling layers. In the case of ABAE we limit the range of the coefficient $\alpha$ between $[0,1]$ introducing a clipping function $\text{clip}(\alpha) = \min(1, \max(0, \alpha))$. 
It is worth noting that, while we consider the method in \cite{huang2017arbitrary}, our framework allows using different style transfer approaches such as the one proposed in \cite{li2017universal}. In this case, the only difference would be the representation of style $s$, which in \cite{li2017universal} is modelled in terms of mean and covariance matrix. 

\begin{algorithm}[t]
 \SetAlgoLined
 \KwData{$\mathbb{O}$: energy function, $M$: number of samples, $\tau$: learning rate}
 \KwResult{Set $\mathcal{S}' = \{z_0, z_1, ..., z_M\}, z_i \sim exp(\mathbb{O})$}
 \tcp{Initialization}
 $z \sim N(0, \mathbb{1})$, $i := 0$, $\mathcal{S}' := \emptyset$\;
 \While{i < M}{
  \tcp{Generate candidate point}
  $\hat{z} := z + l \mathbb{O}_g(z) + \sqrt{2 \tau} \epsilon$\;
  \tcp{Calculate acceptance ratio}
  $r := \exp \left(\mathbb{O}(\hat{z}) - \mathbb{O}(z) + \frac{\left\Vert \hat{z} - z + \tau \nabla \mathbb{O}(z)   \right\Vert}{4\tau} - \frac{\left\Vert z - \hat{z} + \tau \nabla \mathbb{O}(\hat{z})   \right\Vert}{4\tau}  \right)$\;
  \If {$ \mathbb{U}(0, 1) \leq \min(1,r)$} {
    \tcp{Accept candidate point}
    $\mathcal{S}' \leftarrow \hat{z}$, $i := i + 1$, $z := \hat{z}$\;
   }
 }
 \caption{Langevin MCMC}
 \label{alg:mcmc}
\end{algorithm}

The implementation of the perceptual attribute predictor $P_A$ depends on the considered attribute. 
For memorability we resort on the Memnet model introduced in~\cite{khosla2015understanding}
to allow fair comparison with~\cite{siarohin2017make}. 
As suggested in~\cite{khosla2015understanding}, we consider the HybridCNN network \cite{hybrydcnn} and finetune it on LaMem dataset \cite{khosla2015understanding}.  
Following this protocol, the resulting model $\hat{P}_A(I)$ implements a regressor, \textit{i.e.} $\hat{P}_A(I) \in [-\infty, +\infty]$. To normalize the output scores of the memorability predictor we compute
$P_A(I) = \Sigma(\hat{P}_A(I)) ^ \lambda$, where $\Sigma$ is a sigmoid function and $\lambda$ a user defined parameter.
We follow a similar approach for deriving $P_A(I)$ in the case of scariness. 
We use InceptionV3 network as one of the best general purpose models~\cite{szegedy2016rethinking}. We trained this model on images with their binary labels from the BAM dataset \cite{wilber2017bam} to derive $\hat{P}_A(I)$ and then compute $P_A(I) = \hat{P}_A(I) ^ \lambda$.

In the proposed GAN model the generator $G$ is implemented as a neural network with the following structure: $FC_{128}^{R}$ - $FC_{512}^{R}$ - $FC_{1024}$, where $FC_{P}^{R}$ denotes a fully-connected layer with $P$ output units and Relu activation, while $FC_{P}$ indicates a fully-connected layer with $P$ output units without activation. Similarly, the architecture of the discriminator $D$ is defined as: $FC_{512}^{R}$ - $FC_{256}^{R}$ - $FC_{128}^{R}$ - $FC_{1}$.

\textbf{Style Sampling Methods.} 
In this work we used MCMC sampling 
in order to find the best styles. MCMC is a general method for sampling from a multivariate probability distribution. We define the energy function 
$\mathbb{O}(z) = \log \left( P_A(Z(I, G(z))) N(z; 0, \mathbb{1}) \right)$
and we chose Langevin MCMC \cite{rossky1978brownian} as our sampling method (see Algorithm~\ref{alg:mcmc}). For simplicity here we report the formulas only for BAE. The algorithm is similar for ABAE.
We also experiment with the two other popular MCMC methods: Metropolis Hastings \cite{hastings1970monte} and 
Hamiltonian \cite{neal2011mcmc}. 
The effect of using different MCMC methods for creating new styles is discussed in Section~\ref{sec:res}.
We also introduced two modifications to the traditional methods to help increasing the acceptance rate (line 5 - Algorithm~\ref{alg:mcmc}): 
\begin{itemize}[noitemsep,topsep=0pt]
    \item \textit{Adaptive gradient}: Instead of using $\nabla \mathbb{O}(z)$ we consider it adaptive version, in analogy to Adam~\cite{kingma2014adam}. We found this strategy especially helpful for ABAE, because the gradient for $\alpha$ can be several order of magnitude higher than the gradient for $z$. 
    \item \textit{Adaptive learning rate}: At step 5 in Algorithm~\ref{alg:mcmc} upon rejection we decrease the learning rate $\tau$ (\textit{e.g}. $\tau:=0.9\tau$) while upon acceptance we set $\tau$ to the initial value. This strategy eliminates the need of tuning the learning rate $\tau$ for each image.
\end{itemize}

\section{Experimental Validation}
\label{sec:res}
In this Section we report the results of our experimental evaluation. First, we provide some details on the used datasets and our experimental setup (Section~\ref{sec:res_intro}). Then, in Section~\ref{sec:res} we quantitatively evaluate the performance of our method in enhancing two different perceptual attributes: memorability and scariness. In the case of memorability, we also discuss the effect of using different sampling methods. Finally, we report qualitative results comparing our method with baselines. 
Our code is available online~\cite{githubBAE}.

\subsection{Experimental Setup and Datasets}
\label{sec:res_intro}

\textbf{Datasets.} We considered three datasets in our work. \\
The \textit{DevianArt} dataset  \cite{sartori2015affective} is a collection of 500 abstract art paintings collected from an online social network site, deviantart.com, devoted to user-generated art. The dataset was used in \cite{siarohin2017make} to define the style set. \\
\textit{LaMem} \cite{khosla2015understanding} is 
a collection of 58,741 images annotated with the corresponding memorability score. The scores were collected through an efficient version of the memorability game. We encourage the reader to refer at \cite{khosla2015understanding} for further details. This dataset was also considered in \cite{siarohin2017make}. \\
The \textit{Behance-Artistic-Media (BAM)} dataset \cite{wilber2017bam} is a very large dataset with automatically labeled binary attribute scores.
It comprises about 20 attributes, including emotional attributes like scary, gloomy, happy and peaceful. It contains 14,585 images (with positive or negative labels)
originally crowdsourced from human annotators for the scary attribute.
We were able to download a subset of 11,698 images from this dataset. We use this set to train our scariness predictor.\\
\textbf{Experimental Setup.}
We now provide further details on our experimental setting and implementation. We follow an experimental protocol similar to \cite{siarohin2017make} in order to allow a fair comparison with their work. Note that \cite{siarohin2017make} only focuses on memorability, while our approach deals with arbitrary attributes. \\
\textit{Styles set.} For the style set $\Omega_S$ we considered 500 abstract art images from the DeviantArt \cite{sartori2015affective} dataset. 
While Siarohin \textit{et al.} \cite{siarohin2017make} considered a pre-defined set of styles selecting 100 images from this dataset, our approach by learning a style probability density function can potentially learn from and generate an infinite number of styles.  
As described in Section~\ref{sec:BAE}, we use a GAN model to represent the probability density over the set of styles. The GAN was trained with batch size equal to 64 and for about 100k iterations. All the other hyper-parameters were set as indicated in \cite{gulrajani2017improved}.\\
\textit{Baseline methods.} In the case of memorability, we compare the performance of our method with~\cite{siarohin2017make}.
The code from \cite{siarohin2017make} is available online~\cite{githubMEM}. 
This is the closest work to ours, where a set of only 100 styles is considered for increasing memorability. We also consider an additional baseline method $\mathcal{B}$ which uses the same set of 500 styles of our approach. Specifically, this baseline consists in applying the style transfer method in \cite{huang2017arbitrary} to the given image considering all the style pictures in the style set and then compare the obtained stylized images with those we obtained with our method setting $M=500$.
In the case of scariness, we simply compare with $\mathcal{B}$. \\
\textit{Perceptual Attribute Predictors.} For the target attributes, memorability and scariness, we trained two predictors from two independent set of images. The first predictor, which we denote as the internal predictor $\mathbb{M}$, is used for generating the stylized images and corresponds to $\hat{P}_A$, while the second is employed only for assessing the performance of our method and we call it the external predictor $\mathbb{E}$, indicated as $\hat{P}^{\mathbb{E}}_A$. 
Specifically we use the second predictor to compute the score increases between the original image 
and the stylized images obtained with our method. 
In the case of memorability, we split LaMem into two sets of 22,500 images each and use these sets to learn the two predictors. 
This is exactly the same setup used in \cite{siarohin2017make}. 
We also used the same training parameters.
To verify that these predictors are valid and have performance close to human annotators, following \cite{khosla2015understanding}, we compute the rank correlation and we obtain a value of 0.63 for both models. 
As reported in \cite{khosla2015understanding}, this is close to human performance (0.68).
In the case of scariness, we finetuned InceptionV3  \cite{szegedy2016rethinking} (considering only the two top inception blocks) originally trained on ImageNet using labels from BAM dataset \cite{wilber2017bam}. 
We split the BAM labeled set into two disjoint sets of 5,849 images each and trained two scariness prediction models. \\
\textit{Style Transfer.} As stated above for the style transfer network, we used the recent approach from \cite{huang2017arbitrary}. We considered the pretrained network released by \cite{huang2017arbitrary}. 
For the baseline $\mathcal{B}$ and our method BAE we used $\alpha=0.5$. In the case of adaptive alpha, we used a Gaussian distribution as prior $P(\alpha) = N(\alpha; 0.5,0.25)$ (see Eqn.~\ref{eq:alpha}).\\
\textit{Hyper-parameters.} For the experiment on memorability we set $\tau=1e-1$ and $\lambda=100$, while
for scariness we consider $\tau=1e-2$ and $\lambda=10$. 
In general, we found that the higher is $\lambda$ the higher is the attribute increase, but when $\lambda$ is too high nearly all the candidates points $\hat{z}$ are rejected at step 5 in Algorithm~\ref{alg:mcmc}. So, we set $\lambda$ to the highest value (we try values on a log scale, \text{i.e.} $\lambda \in \{1, 10, 100, 1000\}$) for which this effect is not observed. 
A similar trend is observed for the learning rate $\tau$. If $\tau$ is high, we obtain more diverse styles. Still, when the learning rate is too high almost all candidate points are rejected. Similarly to $\lambda$, we set the initial learning rate $\tau$ to the highest possible value for which we do not observe this effect (we also try values on log scale $\tau \in \{ 1e-3, 1e-2, 1e-1, 1 \}$). 
The problem of choosing the optimal learning rate is partially overcome with the adaptive learning rate strategy described in Section~\ref{sec:implem}. Still, choosing an optimal initial learning rate can greatly improve the overall method speed. \\
\textit{Image test set.} We evaluate the performance of our approach on the same test set as in \cite{siarohin2017make}, which we call $\mathcal{V}$, consisting of 1,001 generic images. 
We used this test set also for the experiment on scariness.

\begin{figure}[t]
    \begin{center}
      \includegraphics[width=0.65\linewidth]{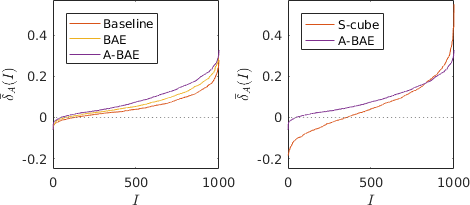}
    \end{center}
    \caption{Sorted memorability differences $\bar{\delta}_A(I)$ 
    for the images $I \in \mathcal{V}$ obtained by averaging over the top 10 results retrieved with each method. Comparison of our methods with (left) the baseline $\mathcal{B}$ and (right) the competing work S-cube~\cite{siarohin2017make}.}
    \label{fig:mem_gaps}
\end{figure}
\begin{figure}[t]
  \begin{center}
     \includegraphics[width=0.7\linewidth]{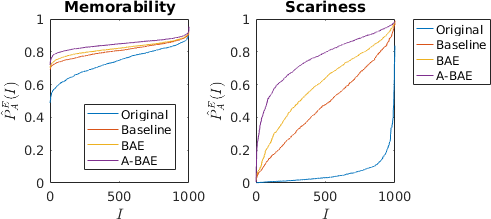} 
     \end{center}
\caption{Perceptual attributes scores. Sorted scores for the original images $I \in \mathcal{V}$ and the top results retrieved in the case of (left) memorability and (right) scariness: original image scores $\hat{P}^\mathbb{E}_A(I)$ and comparison with the top results $\hat{P}^\mathbb{E}_A(\hat{I})$ obtained with the baseline $\mathcal{B}$ and our method.}
    \label{fig:res_scores}
\end{figure}


\begin{table}[t]
\centering
\small
\begin{tabular*}{0.45\linewidth}{c@{\extracolsep{\fill}} cccc} 
\toprule
N  & S-cube \cite{siarohin2017make} & $\mathcal{B}$ & BAE & ABAE \\
\midrule
 1  & 0.0792 & 0.0677 & 0.0812 & \textbf{0.1067} \\
 5  & 0.0594 & 0.0590 & 0.0762 & \textbf{0.0976} \\
 10 & 0.0488 & 0.0544 & 0.0723 & \textbf{0.0911} \\
\bottomrule
 \multicolumn{5}{c}{(a) memorability}
\end{tabular*}
\ \ \ \ \ \
\begin{tabular*}{0.45\linewidth}{c@{\extracolsep{\fill}} ccc}
\toprule
 N & $\mathcal{B}$ & BAE & ABAE \\ \midrule
  1	 & 0.4151 &  0.5362 &  \textbf{0.6960} \\ 
  5	 & 0.3500 &  0.5194 &  \textbf{0.6775} \\ 
 10	 & 0.3153 &  0.5075 &  \textbf{0.6631} \\ 
  \bottomrule
 \multicolumn{4}{c}{(b) scariness}
\end{tabular*}
\caption{Increasing (a) memorability and (b) scariness. Performance of our method with fixed $\alpha$ (BAE) and adaptive $\alpha$ (ABAE) compared to the baseline $\mathcal{B}$ and, in the case of memorability, also to~\cite{siarohin2017make}. Performances are measured in terms of memorability score differences averaged over the top N results $\bar{\delta}_A(I)$.}
\label{tab:perf}
\end{table}

 \subsection{Results}
 \label{sec:res}

\textbf{Evaluation Metrics.} 
Similarly to \cite{siarohin2017make}, we use the \textit{Top N} results and compute the average \textit{score difference} $\bar{\delta}_A(I)$ as evaluation measure. 
Specifically, for each method computing $M$ stylized images, we rank these images based on the attribute scores calculated with the internal predictor $\hat{P}_A(\hat{I})$.
The \textit{Top N} results corresponds to the subset of N images
which rank the highest according to these scores.
Then, given a generic image $I$ and a corresponding stylized image ${T}(I,s)$, we define $\delta_A(I,s)$ as the difference between the attribute scores of these two images, based on the external predictor, \textit{i.e.} $\delta_A(I,s)=\hat{P}_A^\mathbb{E}({T}(I,s))-\hat{P}^\mathbb{E}_{A}(I)$.
Finally, given the Top N results, we compute for each image $I$ the corresponding average score difference $\bar{\delta}_A(I)$ by averaging over $\delta_A(I,s)$ from the Top N set.

\textbf{Quantitative results.} 
We first perform some experiments in order to compare our approach with baseline methods on the memorability enhancement task. 
Figure~\ref{fig:mem_gaps} reports the average memorability differences obtained for all the images $I \in \mathcal{V}$.
In the plot on the left we compare our approach in the case of fixed $\alpha$ (BAE) and adaptive $\alpha$ (ABAE) with the baseline. 
It is straightforward to see that our approach performs better than the baseline $\mathcal{B}$ and that the adaptive method ABAE guarantees a higher memorability gain with respect to BAE using a fixed $\alpha$. 
This indicates that the possibility to automatically set the degree of stylization is beneficial in terms of performance.
In the plot on the right we compare our best performing method ABAE with the competing work \cite{siarohin2017make}.
It can be noted that in the case of \cite{siarohin2017make} the average memorability differences are negative for a large set of test images. 
This difference may be explained by the fact that in the case of \cite{siarohin2017make} the top N styles are retrieved from a pool of only 100 art images, while our method learns the style space from an initial set of 500 styles. This result highlights the importance of considering a wide set of styles, in order to find those which better suit a given image. In this respect, our method is very powerful, being able to interpolate between the styles of a given style set,  thus achieving a higher memorability increase. 
 
 Figure~\ref{fig:res_scores} reports the results of a similar analysis. Specifically, it depicts the sorted scores of the original image set $\mathcal{V}$ and of the corresponding sorted scores of stylized images obtained with our methods and with the baseline $\mathcal{B}$ in the case of (left) memorability and (right) scariness enhancement. 
 For each image $I$ and all the methods we consider only the best stylized image, \textit{i.e.} the one for which we measure the highest attribute score increase. From the figure we can observe that both plots exhibits a similar trend: our method outperforms $\mathcal{B}$ and the adaptive $\alpha$ version ABAE outperforms the fixed $\alpha$ version BAE. It can also be observed that the score increases are more significant in the case of scary images. This result may be partially due to characteristics of the considered dataset: 
 in the case of memorability, most of the images in the dataset exhibit an initial memorability score higher than 0.5, 
 while in the case of scariness the original score is lower than 0.2 for the large majority of the images.
 

\begin{table}[t]
\footnotesize
\begin{center}
\begin{tabular*}{1\linewidth}{c@{\extracolsep{\fill}} cccc}
\toprule
& Top N & M.H. & Langevin & Hamiltonian \\ \midrule
\multirow{3}{*}{BAE} 
 & 1	 & 0.0766 &  \textbf{0.0812} &  0.0780 \\ 
 & 5	 & 0.0725 &  \textbf{0.0762} &  0.0734 \\ 
 & 10	 & 0.0692 &  \textbf{0.0723} &  0.0700 \\ 
\midrule
\multirow{3}{*}{A-BAE} 
 & 1	 & 0.1070 &  0.1067 &  \textbf{0.1094} \\ 
 & 5	 & 0.0995 &  0.0976 &  \textbf{0.1012} \\ 
 & 10	 & 0.0939 &  0.0911 &  \textbf{0.0955}  \\
\bottomrule
\end{tabular*}
\end{center}
\caption{Performance of our methods (top) BAE and (bottom) A-BAE considering different sampling strategies. Performances are measured in terms of memorability score differences averaged over the top N results $\bar{\delta}_A(I)$.}
\label{tab:perf_sampling}
\end{table}

\begin{figure*}[t]
 \begin{center}
 \includegraphics[width=\linewidth]{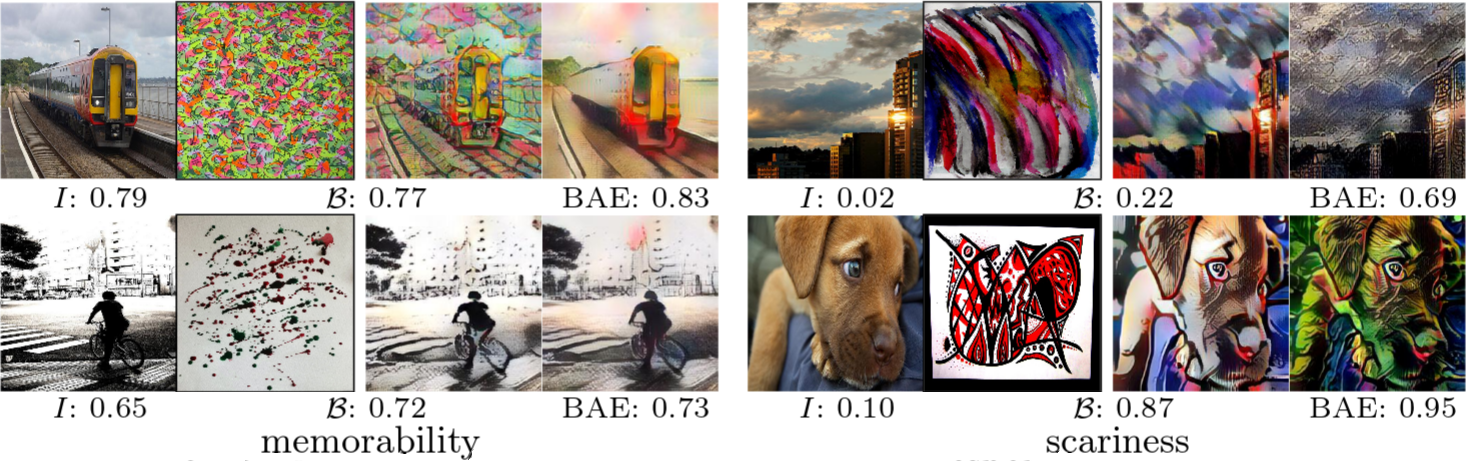}
 \end{center}
\caption{Qualitative results: (left) original image and corresponding top result obtained with (center) the baseline $\mathcal{B}$ and (right) our method BAE. The predicted memorability and scariness scores $\hat{P}^\mathbb{E}_A(\cdot)$ are reported below each image.}
\label{fig:res_tpt}
\end{figure*}
\begin{figure*}[t]

 \begin{center}
 \includegraphics[width=\linewidth]{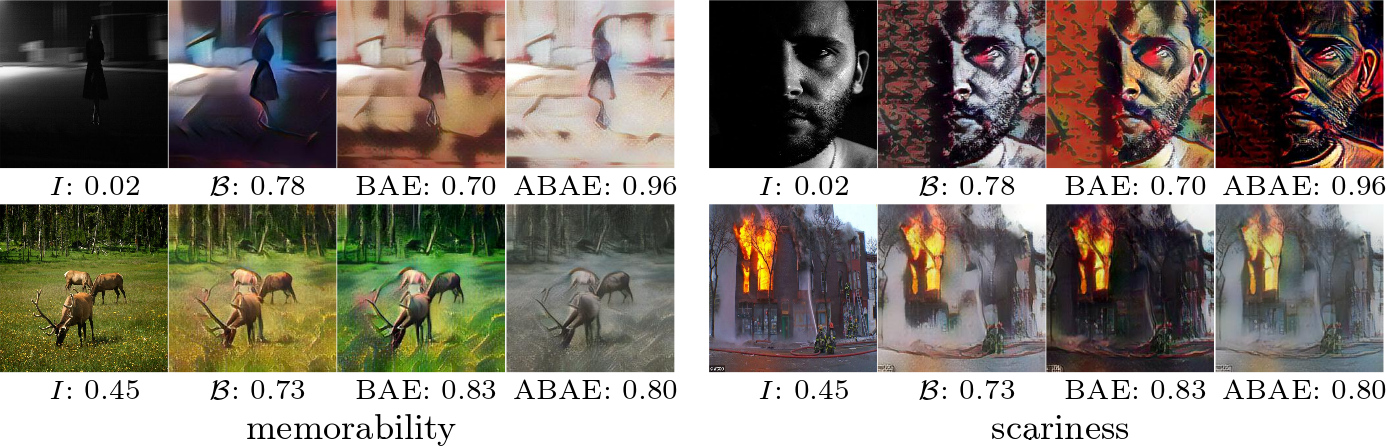}
 \end{center}
\caption{Qualitative results: original input image and top result obtained with the baseline $\mathcal{B}$ and our method with fixed $\alpha$ (BAE) and adaptive $\alpha$ (ABAE). The corresponding memorability and scariness scores $\hat{P}^\mathbb{E}_A(\cdot)$ are reported below each image.}
\label{fig:res_ada_tpt}
\end{figure*}
\begin{figure}[t]
  \scriptsize
\begin{center}
\begin{tabular}{ccccccc}
  \scriptsize
\includegraphics[width=1.8cm,height=1.8cm]{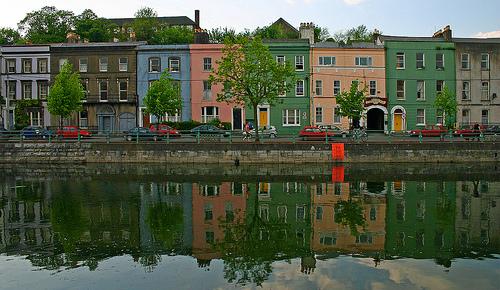} & $\mathcal{B}$: &
\includegraphics[width=1.8cm,height=1.8cm]{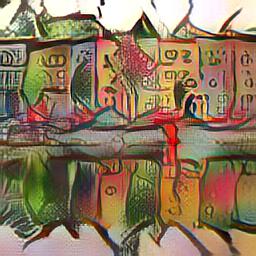} &
\includegraphics[width=1.8cm,height=1.8cm]{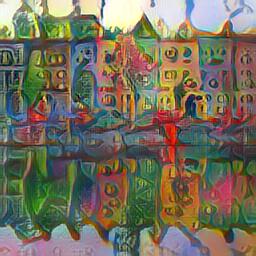} &
\includegraphics[width=1.8cm,height=1.8cm]{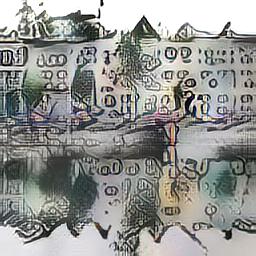} &
\includegraphics[width=1.8cm,height=1.8cm]{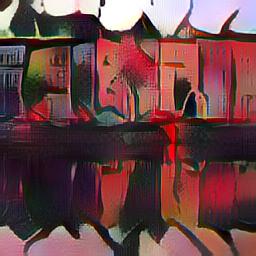} &
\includegraphics[width=1.8cm,height=1.8cm]{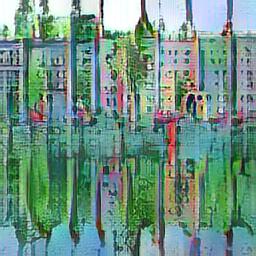} \\
 0.62 & & 1) 0.75 & 2) 0.73 & 3) 0.73 & 4) 0.73 & 5) 0.76 \\ & BAE: &
\includegraphics[width=1.8cm,height=1.8cm]{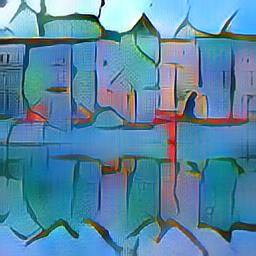} & 
\includegraphics[width=1.8cm,height=1.8cm]{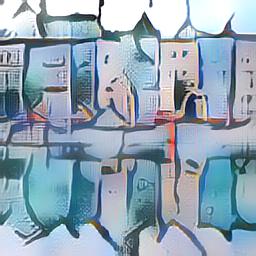} &
\includegraphics[width=1.8cm,height=1.8cm]{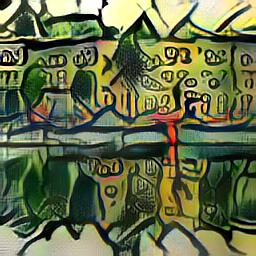} & 
\includegraphics[width=1.8cm,height=1.8cm]{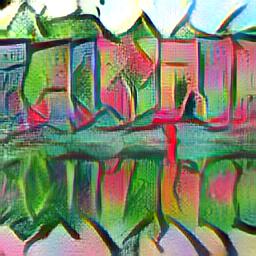} &
\includegraphics[width=1.8cm,height=1.8cm]{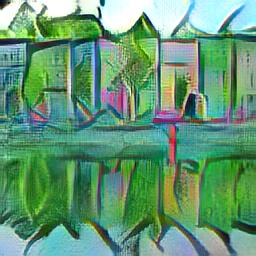} \\
& & 1) 0.80 & 2) 0.78 & 3) 0.80 & 4) 0.75 & 5) 0.77 \\
\includegraphics[width=1.8cm,height=1.8cm]{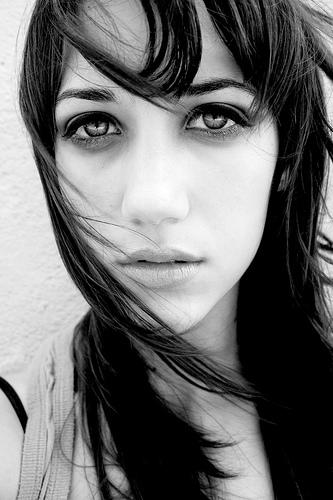} & 
$\mathcal{B}$: &
\includegraphics[width=1.8cm,height=1.8cm]{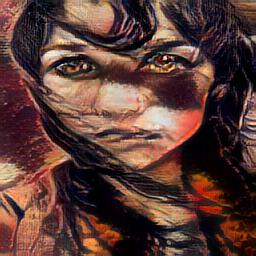} &
\includegraphics[width=1.8cm,height=1.8cm]{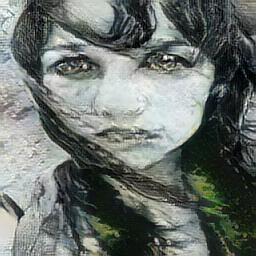} &
\includegraphics[width=1.8cm,height=1.8cm]{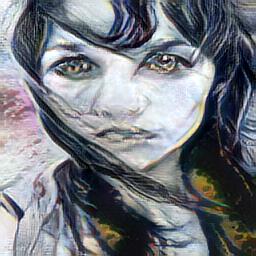} &
\includegraphics[width=1.8cm,height=1.8cm]{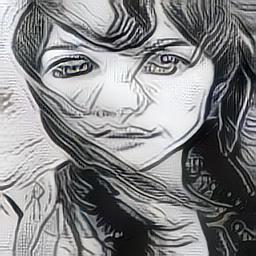} &
\includegraphics[width=1.8cm,height=1.8cm]{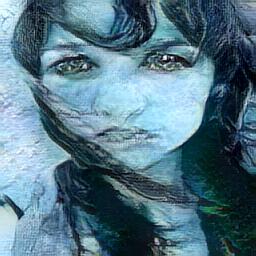} \\
0.01 &  & 1) 0.77 & 2) 0.83 & 3) 0.72 & 4) 0.60 & 5) 0.78 \\ & ABAE: &
\includegraphics[width=1.8cm,height=1.8cm]{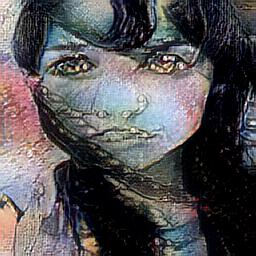} & 
\includegraphics[width=1.8cm,height=1.8cm]{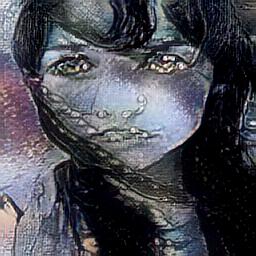} &
\includegraphics[width=1.8cm,height=1.8cm]{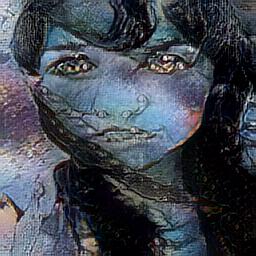} & 
\includegraphics[width=1.8cm,height=1.8cm]{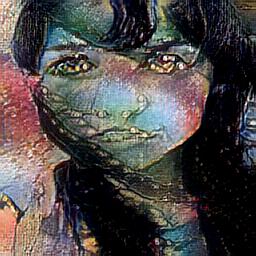} &
\includegraphics[width=1.8cm,height=1.8cm]{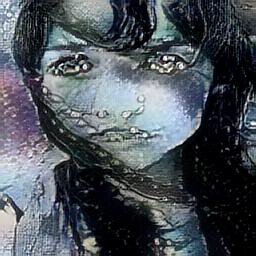} \\
    & & 1) 0.97 & 2) 0.97 & 3) 0.95 & 4) 0.95 & 5) 0.95  
 \end{tabular} 
 \end{center}
\small
\caption{Increasing perceptual attributes: top 5 results for a given sample image. (Left) Original image and (right) comparison (top) in the memorability scenario between  $\mathcal{B}$ and BAE and (bottom) in the scariness scenario between $\mathcal{B}$ and ABAE.}
\label{fig:res_topN}
\end{figure}
  \begin{figure}[t]
  \scriptsize
\begin{center}
\begin{tabular}{cccc}
  \includegraphics[width=1.35cm,height=1.35cm]{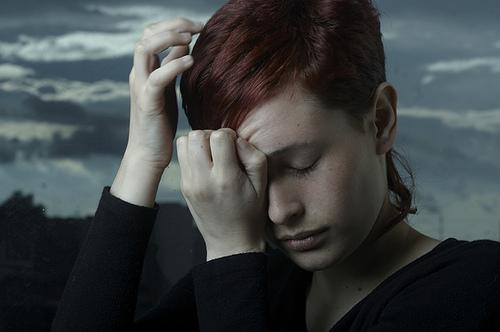} & 
 \includegraphics[width=1.35cm,height=1.35cm]{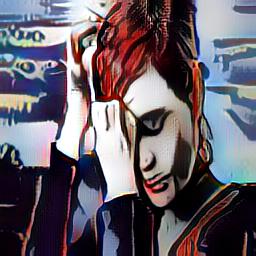}   &  
  \includegraphics[width=1.35cm,height=1.35cm]{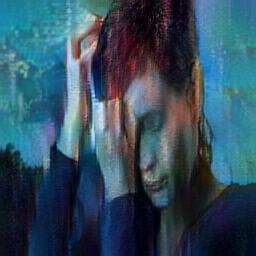} &
 \includegraphics[width=1.35cm,height=1.35cm]{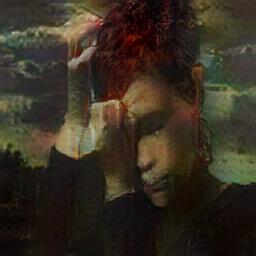} \\
  I$_V$: 0.02  &  $\mathcal{B}$: 0.66 & BAE: 0.45 & ABAE: 0.51 
  \end{tabular}
  \ \ 
\begin{tabular}{cccc}
\includegraphics[width=1.35cm,height=1.35cm]{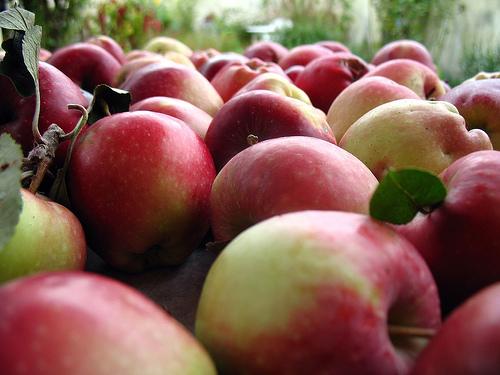} &  
 \includegraphics[width=1.35cm,height=1.35cm]{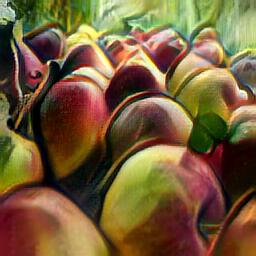}  &  
   \includegraphics[width=1.35cm,height=1.35cm]{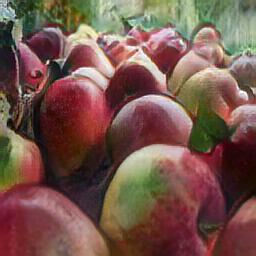} &
 \includegraphics[width=1.35cm,height=1.35cm]{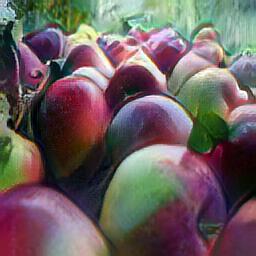} \\
  I$_V$:  0.06  &   $\mathcal{B}$: 0.24 & BAE: 0.21 & ABAE: 0.29 
\end{tabular}
 \end{center}
\small
\caption{Increasing scariness: sample images where our method is not effective. 
}
\label{fig:res_failure}
\end{figure}

 A comparison between our approach and the competing methods is also provided in Table~\ref{tab:perf}(a), where we report the average memorability increases over the Top N results for test set $\mathcal{V}$ in the cases of $N=1$, $N=5$ and $N=10$. 
In all cases our method performs better than the baseline $\mathcal{B}$ and the competing approach \cite{siarohin2017make} and the highest performance is obtained with the adaptive version of our approach. It is also interesting to note that the performances of the baseline $\mathcal{B}$ are sometimes comparable or inferior to those of method in \cite{siarohin2017make}.  Indeed, $\mathcal{B}$ and~\cite{siarohin2017make} are based on two different style transfer approaches, and this may explain the small performance gaps, especially in the case of small N. A similar trend is observed in the experiments on the scariness scenario (Table~\ref{tab:perf}(b)).

\textbf{Comparison between different sampling methods.}
Table~\ref{tab:perf_sampling} reports the results of our approach using different sampling methods in the case of memorability enhancement. 
As expected, Metropolis-Hastings MCMC corresponds to the worst performance, while the other two methods are comparable. Qualitatively, we did not observe significant differences between the three methods. 
In light of these results, in all our experiments we use Langevin MCMC as sampling strategy as it represents the best trade-off between performance and computational speed.

\textbf{Running time.} The running times for the Langevin MCMC of one image on Nvidia Titan X are, respectively, 1m41s for the baseline (500 style images) and 7m20s for A-BAE (500 MCMC iterations). However, by decreasing the number of iterations in A-BAE to 100 the running time can be reduced to 1m28s, while the top1 average difference is still higher than the baseline (0.55 vs 0.41).

\textbf{User study.} We run a user study to show the advantage of using our method for increasing image attribute in the case of scariness. 
The user study consisted in showing pairs of images to a user who was asked to indicate, for each pair, the image which looked more scary. 
We randomly selected 100 images out of our test set, and considered for each image the corresponding top results obtained respectively with the baseline $\mathcal{B}$ and our method ABAE. 
We run the study with 11 people (6 male, 5 female); viewers voted for the image modified with ABAE in the 72.36\% of the cases in average. 
The inter user agreement for this user study, measured with cronbach alpha coefficient is 0.78, thus validating the study.

\textbf{Qualitative results.}  
Finally, we report some qualitative results.
Figure~\ref{fig:res_tpt} depicts sample stylized images obtained with our method and with the baseline $\mathcal{B}$ in the case of (left) memorability and (right) scariness enhancement. Given an input image $I_{\mathcal{V}}$ of the test set $\mathcal{V}$, we report the top stylized image computed by $\mathcal{B}$ and BAE. In both cases, the coefficient $\alpha$ is set to 0.5. In the case of $\mathcal{B}$, we also display the corresponding selected style. For the figure it is interesting to observe that, by generating new styles, our method allows to better customize the style to a given image and to achieve an higher increase in terms of attribute score.
In Figure~\ref{fig:res_ada_tpt} we report additional results to show the effects of further adapting the style to the given image by computing the optimal stylization coefficient $\alpha$. For each image, we report the top result obtained with the baseline $\mathcal{B}$ and with our methods BAE and ABAE. (Due to space limitations, we do not report the original style image for $\mathcal{B}$). Our methods produce a significant increase in terms of perceptual score with respect to $\mathcal{B}$ and generally creates a style which better suit the input image. Furthermore, the performance of ABAE are always close or significantly better than those of BAE. 

So far, we compared the Top 1 results obtained with different methods. 
In Figure~\ref{fig:res_topN} instead we report the top 5 results corresponding to some sample images on the two considered scenarios. Specifically, we show a comparison between (left) $\mathcal{B}$ and BAE in the case of memorability and (right) $\mathcal{B}$ and ABAE in the case of scariness. The result images obtained with our method usually obtain higher score increases. As a counterpart,
these increases come with a small loss in terms of diversity for the top stylized images with respect to $\mathcal{B}$. 
In Figure~\ref{fig:res_failure} we report a few cases where our method does not perform as expected. We report sample results in the scariness scenario. 
In one case (left) our method performs poorly with respect to the baseline. 
In the other case (right) neither the baseline nor our method can find a suitable solution to create a scary picture.

\section{Conclusions}
We presented BAE, a novel framework for generating stylized images in order to enhance a predefined perceptual attribute. By exploiting recent advances on neural style transfer and generative adversarial models, we showed that it is possible to edit images such as to increase their memorability and scariness. 
Future work will be devoted to exploit different style transfer approaches and consider other subjective properties.

\section{Acknowledgments}
We gratefully acknowledge Fondazione Caritro for supporting SMARTourism project 
and NVIDIA Corporation for the donation of the TitanX GPU used for this research.

\bibliographystyle{splncs04}


%





\end{document}